\pdfoutput=1
\documentclass[11pt]{article}

\usepackage[]{acl}

\usepackage{times}
\usepackage{latexsym}

\usepackage[T1]{fontenc}
\usepackage[utf8]{inputenc}

\usepackage{microtype}

\usepackage{soul}
\usepackage{url}
\usepackage[utf8]{inputenc}
\usepackage{caption}
\usepackage{graphicx}
\usepackage{amsmath}
\usepackage{amsthm}
\usepackage{booktabs}
\usepackage{algorithm}
\usepackage{algorithmic}
\usepackage{pifont}
\usepackage{amssymb}
\usepackage{graphicx}
\usepackage{graphics}
\usepackage{hyperref}
\usepackage{url}
\usepackage{multirow}
\usepackage{colortbl}
\usepackage{makecell}

\usepackage{booktabs}
\usepackage{arydshln}

\usepackage{enumitem}
\usepackage[list=true]{subcaption}
\usepackage[resetlabels]{multibib}
\newcites{sec}{References}

\def\best{\bf\cellcolor[gray]{0.85}}
\def\secbest{\cellcolor[gray]{0.92}}
\definecolor{mediumelectricblue}{rgb}{0.01, 0.31, 0.59}

\newcommand{\electricblue}[1]{\textcolor{mediumelectricblue}{#1}}

\newcommand{\method}{\textsc{LaSS}}
\newcommand{\texthrt}[1]{\textsl{#1}}
\newcommand{\textspt}[1]{\texttt{#1}}
\newcommand{\correctmark}{\textcolor{green}{\ding{51}}}
\newcommand{\wrongmark}{\textcolor{red}{\ding{55}}}
\newcommand{\comm}[1]{}

\title{Joint Language Semantic and Structure Embedding for \\Knowledge Graph Completion}

\author{Jianhao Shen\textsuperscript{1},  Chenguang Wang\textsuperscript{2}\thanks{~~Corresponding author} , Linyuan Gong\textsuperscript{3}, Dawn Song\textsuperscript{3}\\
\textsuperscript{1}Peking University, \textsuperscript{2}Washington University in St. Louis, \textsuperscript{3}UC Berkeley\\\texttt{jhshen@pku.edu.cn}, \texttt{chenguangwang@wustl.edu}, \\\texttt{\{gly,dawnsong\}@berkeley.edu}}

\begin{document}
\maketitle

\begin{abstract}
The task of completing knowledge triplets has broad downstream applications. Both structural and semantic information plays an important role in knowledge graph completion. Unlike previous approaches that rely on either the structures or semantics of the knowledge graphs, we propose to jointly embed the semantics in the natural language description of the knowledge triplets with their structure information. Our method embeds knowledge graphs for the completion task via fine-tuning pre-trained language models with respect to a probabilistic structured loss, where the forward pass of the language models captures semantics and the loss reconstructs structures. Our extensive experiments on a variety of knowledge graph benchmarks have demonstrated the state-of-the-art performance of our method. We also show that our method can significantly improve the performance in a low-resource regime, thanks to the better use of semantics. The code and datasets are available at \url{https://github.com/pkusjh/LASS}.
\end{abstract}
\section{Introduction}
Knowledge graphs (KG), such as Wikidata and Freebase~\cite{bollacker_freebase:_2008}, consist of factual triplets. KGs have been useful resources for both humans and machines. A triplet in the form of \texthrt{(head entity, relation, tail entity)}, where the relation involves both head and tail entities, has been used in a great variety of applications, such as question answering~\cite{Guu2015TraversingKG,hao_end--end_2017} and web search~\cite{Xiong2017ExplicitSR}. Incompleteness has been a longstanding issue in KGs~\cite{Carlson2010TowardAA}, impeding their wider adoption in real-world applications.

KG completion aims to predict a missing entity or relation of a factual triplet.
Structural patterns in the existing triplets are useful to predict the missing elements~\cite{TransE,RotateE}. For example, a composition pattern can be learned to predict the relation \texthrt{grandmother\_Of} based on two consecutive \texthrt{mother\_Of} relations. Besides the structure information, semantic relatedness between entities and relations is also critical to infer entities or relations with similar meanings~\cite{KB_NL_6,KGBERT,wang_structure-augmented_2021}. For example, if a relationship \texthrt{CEO\_Of} holds between two entities, the relation \texthrt{employee\_Of} also holds. There are two kinds of KG completion approaches that fall into different learning paradigms. First, the structure-based approaches treat entities and relations as nodes and edges, and use graph embedding methods to learn their representations. Second, the semantic-based approaches encode the text description of entities and relations via language models. While both structures and semantics are vital to KG completion, it is non-trivial for existing methods to process both structural and semantic information.

\begin{figure*}[h]
    \centering
    \includegraphics[width=0.70\textwidth]{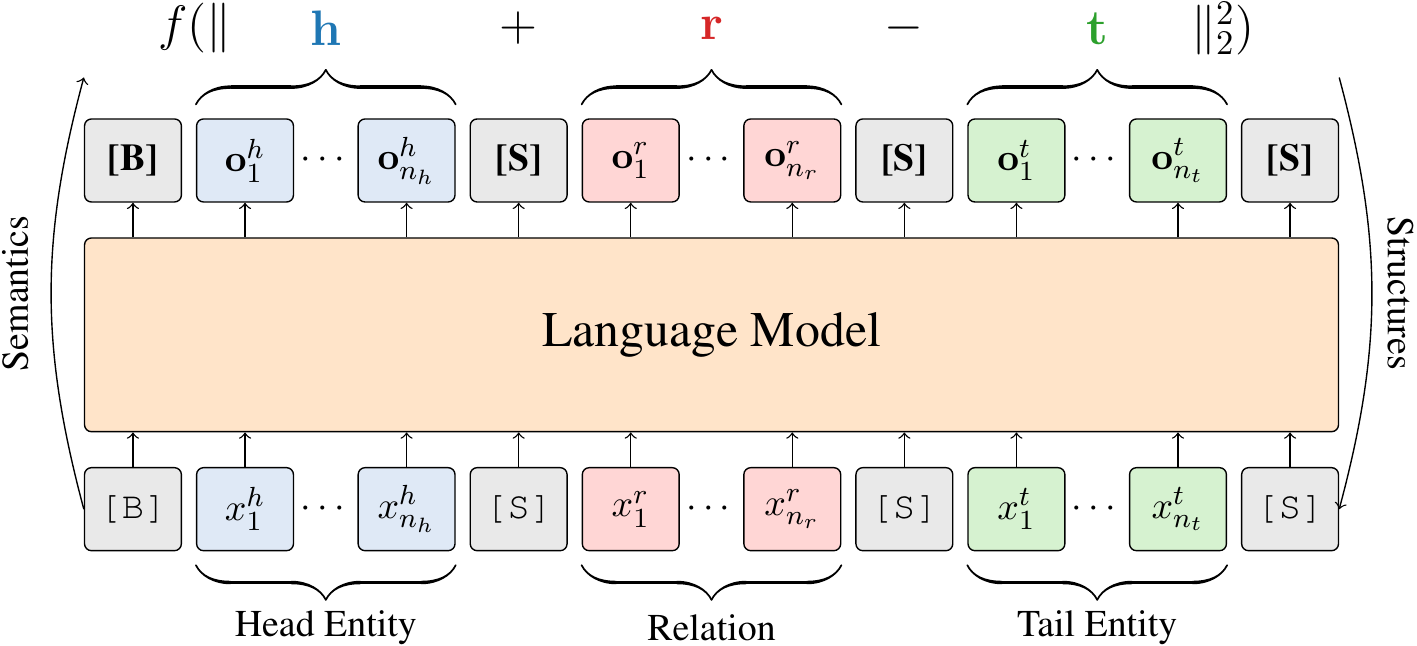}
    \caption{{\small Overview of \method. \method\ maps a knowledge triplet \texthrt{(Head Entity, Relation, Tail Entity)}, in short \texthrt{(h, r, t)}, to the corresponding embedding vectors, $\mathbf{h}, \mathbf{r}, \mathbf{t} \in \mathbb{R}^k$. \method\ embeds KGs for KG completion via fine-tuning pre-trained language models (LM) w.r.t. a probabilistic structured loss, where the forward pass of the LMs captures semantics and the loss reconstructs structures. In particular, \method\ consists of semantic embedding and structure embedding. The {\em semantic embedding} (leftmost arrow) is generated by a forward pass of the LMs followed by a pooling layer over the natural language description of a triplet. \textspt{[B]} (the beginning token) and \textspt{[S]} (the separator token) are special tokens of LMs attached to the description. For example, the textual description of head entity is $(x_1^h, \cdots, x_{n_h}^h)$. $\mathbf{h}$ is calculated as the mean pooling of the corresponding LM outputs $(\mathbf{o}_1^h, \cdots, \mathbf{o}_{n_h}^h)$. $\mathbf{r}$ and $\mathbf{t}$ are calculated similarly. The {\em structure embedding} (rightmost arrow) reconstructs KG structures in the semantic embeddings via optimizing a structured loss on top of the LMs through backpropagation. The structured loss is based on a score function $f(\Vert \mathbf{h}+\mathbf{r}-\mathbf{t} \Vert_2^2)$, which regards the relationship between two entities corresponds to a translation between the embeddings of the entities. The goal is to minimize the loss function so that $\mathbf{h} + \mathbf{r} \approx \mathbf{t}$ when \texthrt{(h, r, t)} holds.}
      \label{fig:overview}}
\end{figure*}

In this paper, we propose \method, a joint language semantic and structure embedding for knowledge graph completion, which incorporates both semantics and structures in a KG triplet. \method\ embeds a triplet into a vector space by fine-tuning pre-trained language models (LM) with respect to a structured loss. \method\ involves both semantic embedding and structure embedding. The semantic embedding captures the semantics of the triplet, which corresponds to the forward pass of a pre-trained LM over the natural language description of the triplet. The structure embedding aims to reconstruct the structures in the semantic embedding, which corresponds to optimizing a probabilistic structured loss via the backpropagation of the LM. Intuitively, the structured loss treats the relationship between two entities as a translation between embeddings of the entities. 
\method\ outperforms the existing approaches on a collection of KG completion benchmarks. We further evaluate \method\ in low-resource settings and find that it is more data-efficient than other methods. The reason is that our method exploits both semantics and structures in the training data.

The contributions are the following: 
\begin{itemize}
    \item We design a natural language embedding approach, \method, that integrates both structural and semantic information of KGs, for KG completion. We train \method~by fine-tuning pre-trained LMs w.r.t. a structured loss, where the forward pass of the LMs captures semantics and the loss reconstructs structures. The method consists of both the KG module and the LM module, which sheds light on the connections between the KGs and deep language representation, and advances the research at the intersection of the two areas.
    \item We evaluate \method~on two KG completion tasks, link prediction and triplet classification, and obtain state-of-the-art performance. The results suggest that capturing both semantics and structures is critical to understand the KGs. The findings are beneficial to many downstream knowledge-driven applications.
    \item We show that we can significantly improve the performance in the low-resource settings over existing approaches, thanks to the improved transfer of knowledge about semantics.
\end{itemize}

\section{\method}
We introduce \method~to embed both semantics and structures of knowledge graphs (KG) with natural language. As shown in Figure~\ref{fig:overview}, \method~incorporates two embeddings: semantic embedding and structure embedding. The semantic embedding captures the semantics in the natural language description of the KG triplets. The structure embedding further reconstructs the structure information of the KGs from the semantic embedding. \method~embeds KG in a vector space by fine-tuning a pre-trained language model (LM) w.r.t. a structured loss, where the forward pass performs semantic embedding and the optimization of structured loss conducts structure embedding.

\subsection{Semantic Embedding}
\label{sec:semantic}

A KG of triplets is denoted as $G$. Each triplet of $G$ is in the form of \texthrt{(h, r, t)}, where \texthrt{h,t} $\in E$ and \texthrt{r} $\in R$. $E$ is the set of entities, and $R$ is the set of relations. The semantic similarities between the head entity \texthrt{h}, relation \texthrt{r}, and tail entity \texthrt{t} are crucial to complete a factual triplet. For example, given \texthrt{h} = ``Bob Dylan'' and \texthrt{r} = ``was born in'', the task is to predict a missing \texthrt{t}, where the candidates are ``Duluth'' and ``Apple''. The semantic similarity between ``Bob Dylan'' and ``Duluth'', as well as the similarity between ``was born in'' and ``Duluth'' should be larger than their similarities with ``Apple'' as ``Duluth'' is the ground-truth answer. Pre-trained LMs capture the rich semantics in natural language via pre-training on large-scale textual corpora. This inspires us to use the semantics stored in the parameters of LMs to encode the semantics of triplets. 

Formally, for a triplet \texthrt{(h, r, t)}, both entities (\texthrt{h} and \texthrt{t}) and relation (\texthrt{r}) are represented by their corresponding natural language descriptions. The head entity \texthrt{h} is represented as a sequence of tokens, $T^h = (x_1^h, \cdots, x_{n_h}^h)$, describing the entity.
Similarly, $T^t = (x_1^t, \cdots, x_{n_t}^t)$ represents the tail entity \texthrt{t}. $T^r = (x_1^r, \cdots, x_{n_r}^r)$ denotes the relation \texthrt{r}.
We generate the semantic embedding via the forward pass of the LMs as shown in Figure~\ref{fig:overview}. The knowledge graph completion tasks require explicit modeling of dependency of the head, relation and tail. For example, both the connections between head and tail, and relation and tail contribute to the prediction of the tail in the link prediction task. Therefore, we use the concatenation of $T^h$, $T^r$, and $T^t$ as the input sequence to the LMs, and use the mean pooling over the output representation of every token in $T^h$, $T^r$, and $T^t$ from the forward pass of LMs as $\mathbf{h}, \mathbf{r}, \mathbf{t}\in \mathbb{R}^k$, where $k$ is the dimension of the embedding vectors. 

More specifically, we construct the input sequence in the following format: \textspt{[B]} $T^h$ \textspt{[S]} $T^r$ \textspt{[S]} $T^t$ \textspt{[S]}, where \textspt{[B]} is a special symbol added in front of every input sequence, and \textspt{[S]} is a special separator token. The special tokens are different for various LMs. For example, \textspt{[B]} and \textspt{[S]} are implemented as \textspt{[CLS]} and \textspt{[SEP]} for BERT~\cite{Devlin_Chang_Lee_Toutanova_2019} respectively. The input sequence is then converted to the corresponding input embeddings of the LMs. For example, the input embeddings of BERT are the sum of the token embeddings, the segment embeddings, and the position embeddings. The input embeddings are fed into the LM. We add a mean pooling layer on top of the output layer of the LM and perform mean pooling over the output representation of every token in $T^h$, i.e., $(\mathbf{o}_1^h, \cdots, \mathbf{o}_{n_h}^h)$, resulting in $\mathbf{h}$ as illustrated in Figure~\ref{fig:overview}. We obtain $\mathbf{r}$ and $\mathbf{t}$ in the same way. The dimension $k$ equals to the hidden size of the LM.

\subsection{Structure Embedding}
Structural information of KGs has been successfully used in the KG completion. Traditional approaches regard the relationship between two entities corresponds to a translation between the embeddings of the entities. This is different from the above semantic embedding and the forward pass cannot capture the structure information. We propose to incorporate the structure embedding by fine-tuning the pre-trained LM with a structure loss.

The goal is to reconstruct structure information in the semantic embedding. The updated embeddings of \texthrt{h}, \texthrt{r}, and \texthrt{t} are still denoted as $\mathbf{h}$, $\mathbf{r}$, and $\mathbf{t}$, which incorporate structure information of KGs while preserve semantic information. We reconstruct structure information in the semantic embeddings via optimizing a probabilistic structured loss, in which the score function of a triplet \texthrt{(h, r, t)} is defined by Eq.~\ref{eq:score}:

\begin{equation}\label{eq:score}
    f(\mathbf{h},\mathbf{r},\mathbf{t}) = b - \frac{1}{2} \Vert \mathbf{h}+\mathbf{r}-\mathbf{t} \Vert_2^2
\end{equation}

If \texthrt{(h, r, t)} holds, we have $\mathbf{h} + \mathbf{r} \approx \mathbf{t}$. We also use $f(\Vert \mathbf{h}+\mathbf{r}-\mathbf{t} \Vert_2^2)$ to denote this in Figure~\ref{fig:overview} for simplicity. The score function is motivated by TransE~\cite{TransE}.

We define the following probabilistic model based on the score function (\ref{eq:score}):
\begin{equation}\label{eq:softmax}
    \Pr({h}|{r},{t})  = \frac{\text{exp}(f(\mathbf{h},\mathbf{r},\mathbf{t}))}{\sum_{\tilde{h} \in E} \text{exp}(f(\tilde{\mathbf{h}},\mathbf{r},\mathbf{t}))}
\end{equation}
Here $\tilde{h}$ is the corrupted head sampled from the entity set $E$. $\Pr(r|h,t)$ and $\Pr(t|h,r)$ have a similar form except that the summation in the denominator is over corrupted relations and tails, respectively. 

The probabilistic structured loss is defined in Eq.~\ref{eq:know_loss}. The goal is to minimize the negative log likelihood over the KG:
\begin{equation}\label{eq:know_loss}
\begin{aligned}
    L = - \sum_{{(h, r, t)} \in G} (& \log \Pr({h}|{r},{t}) + \log \Pr({r}|{h},{t}) \\ 
    & + \log \Pr({t}|{h},{r}))
\end{aligned}
\end{equation}

\paragraph{Optimization}
Computing the probability in Eq.~\ref{eq:softmax} is computationally inefficient since it requires a forward pass of all possible triplets $(\tilde{h}, r, t)$ to compute the denominator. We use negative sampling~\cite{Mikolov2013DistributedRO} to make training more efficient. Instead of minimizing $-\log \Pr(h|r,t)$ as in Eq.~\ref{eq:know_loss}, we optimize the loss as is described in Eq.~\ref{eq:ns} for modeling \texthrt{h}. 

\begin{equation}\label{eq:ns}
\begin{aligned}
    L_\texthrt{h} = - &\log\Pr(1|h, r, t) \\
    - &\sum_{i}^{n_\text{ns}} \mathbb{E}_{\tilde{h}_i\sim E \backslash \{h\}}\log \Pr(0 | \tilde{h}_i, r, t)
\end{aligned}
\end{equation}

where $\Pr(1|h, r, t) = \sigma(f(\mathbf{h}, \mathbf{r}, \mathbf{t}))$.

The loss for modeling \texthrt{r} and \texthrt{t} is similarly defined. Here, hyperparameter $n_\text{ns}$ is the number of negative samples. Each negatively sampled head $\tilde{h}_i$ is drawn uniformly without replacement from the entity set $E \backslash \{h\}$. A sample is not treated as a negative sample if it is already a positive example. 
We have the final structured loss $L=\sum_{{(h, r, t)} \in G} (L_\texthrt{h} + L_\texthrt{r} + L_\texthrt{t})$ by adopting the similar negative sampling procedures for relations and tail entities.

The training of \method\ is unified as fine-tuning an LM with respect to a structured loss. The semantic embedding is obtained by the forward pass of the LM. The structure embedding is conducted by optimizing the structured loss through backpropagation of the LM.

\section{Experiments}
\label{sec:exp}

\subsection{Experimental Setup}
\paragraph{Datasets} We test the performance of our method on five KG benchmarks built with three KGs: Freebase~\cite{bollacker_freebase:_2008}, WordNet~\cite{miller_wordnet:_1995} and UMLS~\cite{ConvE}. Freebase is a large-scale KG containing general knowledge facts. We employ two subsets from Freebase, namely FB15K-237~\cite{toutanova_observed_2015}, and FB13~\cite{KB_NL_1}. WordNet provides semantic knowledge of words. We use two subsets from WordNet, namely WN18RR~\cite{ConvE}, and WN11~\cite{KB_NL_1}. 
UMLS is a medical semantic network containing semantic entities and relations. The statistics are summarized in Table~\ref{tab:statistics}. We also provide a detailed description of the datasets in Appendix~\ref{apx:data}.

\begin{table}[htbp]
\centering
\resizebox{0.95\linewidth}{!}
  {
    \begin{tabular}{l|c|c|c|c|c}
        \toprule
        {\bf Dataset} & $\#$ {\bf Entity} & $\#$ {\bf Relation} & $\#$ {\bf Train} & $\#$ {\bf Dev} & $\#$ {\bf Test}\\
        \hline
        FB15k-237 &  14,541 & 237 & 272,115 & 17,535 & 20,466 \\
        \hline
        % FB15K & 14,951 & 1,345 & 483,142 & 50,000 & 59,071 \\
        % \hline
        WN18RR & 40,943 & 11 & 86,835 & 3,034 & 3,134 \\
        \hline
        % WN18 & 40,943 & 18 & 141,442 & 5,000 & 5,000 \\
        % \hline
        UMLS & 135 & 46 & 5,216 & 652 & 661 \\
        \hline
        FB13 & 75,043 & 13 & 316,232 & 5,908 & 23,733 \\
        \hline
        WN11 & 38,696 & 11 & 112,581 & 2,609 &  10,544 \\
        \bottomrule
    \end{tabular}}
    \caption{{\small Statistics of knowledge graphs.}}
    \label{tab:statistics}
\end{table}
\paragraph{Implementation Details} We use two families of LMs with \method. First, we adopt both BERT$_{\rm BASE}$ and BERT$_{\rm LARGE}$ from ~\cite{Devlin_Chang_Lee_Toutanova_2019} with \method, namely \method-BERT$_{\rm BASE}$ and \method-BERT$_{\rm LARGE}$. Second, RoBERTa family~\cite{Liu_Ott_Goyal_Du_Joshi_Chen_Levy_Lewis_Zettlemoyer_Stoyanov_2019} is used, namely \method-RoBERTa$_{\rm BASE}$ and \method-RoBERTa$_{\rm LARGE}$.

We train \method\ with AdamW~\cite{loshchilov_decoupled_2018} on each KG dataset via fine-tuning the corresponding LMs. The training hyperparameters are set as follows. For \method-BERT$_{\rm BASE}$ and \method-RoBERTa$_{\rm BASE}$, the batch size is set to 128, the learning rate is set to 3e-5 with linear warm-up and 0.01 weight decay. We set the batch size to 64 for \method-BERT$_{\rm LARGE}$ and \method-RoBERTa$_{\rm LARGE}$. The number of training epochs is set to 5. The margin $b$ in Eq.~\ref{eq:score} is empirically set to 7. We sample 5 negative entities or relations resulting in 15 negative triplets for each positive triplet for the negative sampling. 

We represent entities and relations as their names or descriptions~\cite{KGBERT}. For FB15k-237, we used entity descriptions from ~\cite{KB_NL_3}. For FB13, we use entity descriptions in
Wikipedia. For WN18RR, we use definitions of synsets as entity descriptions. For WN11 and UMLS, the entity names are used as the entity descriptions. The relation descriptions are based on the relation names across all the datasets. The input sequence is constructed based on Sec.~\ref{sec:semantic}.
For \method-BERT$_{\rm BASE}$ and \method-BERT$_{\rm LARGE}$, we use a character-level BPE vocabulary. \textspt{[B]} is replaced  with \textspt{[CLS]}, and \textspt{[S]} is replaced with \textspt{[SEP]}. For \method-RoBERTa$_{\rm BASE}$ and \method-RoBERTa$_{\rm LARGE}$, we use a byte-level BPE vocabulary, and \textspt{[B]} and \textspt{[S]} are replaced with \textspt{BOS} and \textspt{EOS} respectively. We implement \method\ using the Transformers package~\cite{Wolf2019HuggingFacesTS}.

\paragraph{Comparison Methods} We compare our method to state-of-the-art methods, including (\expandafter{\romannumeral1}) shallow structure embedding: TransE~\cite{TransE}, TransH~\cite{TransH}, TransR~\cite{TransR}, TransD~\cite{ji_knowledge_2015}, TransG~\cite{xiao_transg_2016}, TranSparse~\cite{ji_knowledge_2016}, DistMult~\cite{DistMult}, DistMult-HRS~\cite{zhang_knowledge_2018}, ConvE~\cite{ConvE}, ConvKB~\cite{ConvKB}, ComplEx~\cite{trouillon_complex_2016}, RotatE~\cite{RotateE},
REFE~\cite{chami-etal-2020-low},
HAKE~\cite{zhang_learning_2019},
and ComplEx-DURA~\cite{NEURIPS2020_f6185f0e};
(\expandafter{\romannumeral2}) deep structure embedding: NTN~\cite{KB_NL_1}, DOLORES~\cite{wang_dolores:_2018}, KBGAT~\cite{nathani-etal-2019-learning}, GAATs~\cite{wang_knowledge_2020}, NePTuNe~\cite{abs-2104-07824}, and ComplEx-N3-RP~\cite{ChenM0S21};
(\expandafter{\romannumeral3}) language semantic embedding: TEKE~\cite{wang_text-enhanced_2016}, KG-BERT~\cite{KGBERT}, and stAR~\cite{wang_structure-augmented_2021}.
We present a detailed technical description of the above methods in Appendix~\ref{apx:comp}.

\begin{table}[htbp]
\centering
\resizebox{1.0\linewidth}{!}
  {
    \begin{tabular}{l|cc|c}
       \toprule
   {\bf Method} & {\bf WN11} & {\bf FB13} & {\bf Avg} \\ 
    \hline
    NTN~\cite{KB_NL_1} & 86.2 & 90.0 & 88.1 \\
    TransE~\cite{TransE} & 75.9  & 81.5  & 78.7 \\
    TransH~\cite{TransH} & 78.8 & 83.3 & 81.1 \\
    TransR~\cite{TransR} & 85.9 &82.5  & 84.2 \\
    TransD~\cite{ji_knowledge_2015} &86.4  & 89.1 & 87.8 \\
    TEKE~\cite{wang_text-enhanced_2016} &86.1  &84.2  & 85.2  \\
    TransG~\cite{xiao_transg_2016} &87.4  &87.3  &87.4  \\
    TranSparse-S~\cite{ji_knowledge_2016} &86.4  &88.2  &87.3  \\
    DistMult~\cite{DistMult} &87.1  &86.2  &86.7  \\
    DistMult-HRS~\cite{zhang_knowledge_2018} &88.9  &89.0  &89.0  \\
    AATE~\cite{KB_NL_6} &88.0  &87.2  &87.6  \\
    ConvKB~\cite{ConvKB} &87.6  &88.8  &88.2  \\
    DOLORES~\cite{wang_dolores:_2018} &87.5  & 89.3   &88.4  \\
    KG-BERT~\cite{KGBERT} & 93.5  & 90.4   & 91.9  \\
    \hdashline
    \method-BERT$_{\rm BASE}$ (\electricblue{\small ours}) & 93.3 &   91.2 & 92.3 \\
    % \method-BERT$_{\rm BASE}$ (0.1) & 88.9 & 90.2  &  \\
    % \method-BERT$_{\rm BASE}$ (0.2) & 90.2 &   &  \\
    % \method-BERT$_{\rm BASE}$ (0.3) &  &   &  \\
    % \method-RoBERTa$_{\rm BASE}$ (0.1) &  &   &  \\
    % \method-RoBERTa$_{\rm BASE}$ (0.2) &  &   &  \\
    % \method-RoBERTa$_{\rm BASE}$ (0.3) &  &   &  \\
    \method-BERT$_{\rm LARGE}$ (\electricblue{\small ours}) & \best 94.5 & \best 91.8 & \best 93.2 \\
    \method-RoBERTa$_{\rm BASE}$ (\electricblue{\small ours}) & 92.3 & 91.1 & 91.7 \\
    \method-RoBERTa$_{\rm LARGE}$ (\electricblue{\small ours}) & \secbest 93.8 & \secbest 91.6 & \secbest 92.7 \\
    % \method-RoBERTa$_{\rm LARGE}$ (\electricblue{\small ours}) & \secbest 93.8 & 91.14 & 92.5 \\
    \bottomrule
    \end{tabular}}
    \caption{{\small Triplet classification accuracy on WN11 and FB13.}}
    \label{tab:tc_results}
    \vspace{-0.2in}
\end{table}

\subsection{Triplet Classification}
The task of triplet classification judges whether a given triplet \texthrt{(h, r, t)} is correct or not. The task is a binary classification task. We use WN11 and FB13 for the task, since only the test sets of the two datasets contain both positive and negative triplets among all the datasets. For the task, we use the score function as defined in Eq.~\ref{eq:score} , and set a score threshold. For a triplet, if the score is above the threshold, the triplet is classified as positive, otherwise negative. We set the threshold empirically based on the accuracy on the validation set. 
As shown in Table~\ref{tab:tc_results}, we conclude with the following findings.
\begin{table}[htb]
    \centering
    \resizebox{0.95\linewidth}{!}{
    \begin{tabular}{cccc}
     \toprule
    Head & Relation & Tail & Label \\
    \midrule
    ron ziegler & gender & male & \correctmark \\
    john fortescue & profession & writer & \correctmark \\
    george j adams & cause of death & typhoid fever & \correctmark \\
    fleiss joseph & institution & columbia university & \correctmark \\
    edmund husserl & nationality & austria & \correctmark \\
    aleksandr bakulev & gender & female & \wrongmark \\
    emile littre & profession & physicist & \wrongmark \\
    joseph smith jr & cause of death & emphysema & \wrongmark \\
    frank g slaughter & institution & university of toronto & \wrongmark \\
    julius klinger & nationality & romania & \wrongmark \\
    \bottomrule
    \end{tabular}}
    \caption{\small {Samples of \method's correct predictions on FB13, where KG-BERT~\cite{KGBERT} outputs wrong predictions. Label \correctmark\ means a gold positive triplet. \wrongmark\ indicates a gold negative triplet.}}
    \label{tab:good_case_fb13}
\end{table}

\begin{figure}[htbp]
    \centering
    \includegraphics[width=0.45\textwidth]{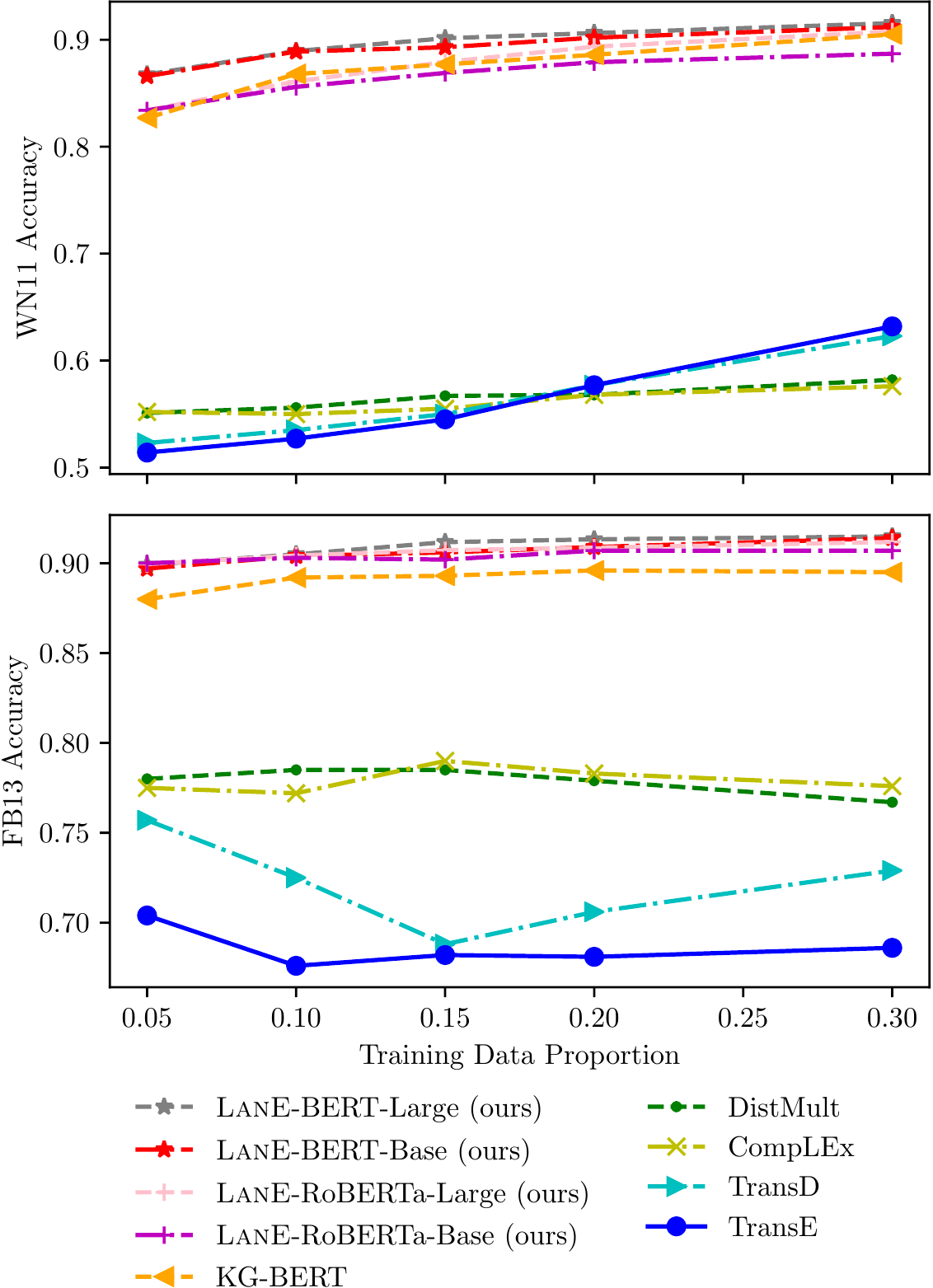}
    \caption{{\small Triplet classification accuracy in a low-resource regime: training with different proportions of the corresponding training datasets on WN11 and FB13.}}
    \label{fig:partial-bert}
\end{figure}

We find that our methods consistently produce state-of-the-art results on triplet classification tasks. This indicates that our score function has captured semantics and structures that are crucial for the triplet classification. We also notice that \method-BERT generates slightly better results compared to \method-RoBERTa. This is due to RoBERTa removing the NSP objective, however the objective naturally fits in the triplet classification task. \method-RoBERTa still generates reasonable results. The reason is that the masked LM objective captures the necessary semantics needed for the triplet classification, and \method\ is able to preserve the important semantic information. 

In Table~\ref{tab:good_case_fb13}, we also show some cases where \method-BERT$_{\rm BASE}$ makes correct predictions while KG-BERT produces incorrect ones on FB13. Compared to KG-BERT, we find that \method\ is more capable in relations that require comprehensive structure information, such as ``institution''.

\begin{table*}[t]
\centering
\resizebox{0.85\linewidth}{!}
  {
    \begin{tabular}{l|cc|cc|cc}
    \toprule
   \multirow{2}{*}{{\bf Method}} & \multicolumn{2}{c|}{{\bf FB15k-237}}  &\multicolumn{2}{c|}{{\bf WN18RR}} &  \multicolumn{2}{c}{{\bf UMLS}}\\
    % \cline{2-5}
     & {\bf Hits@10} & {\bf MR} &  {\bf Hits@10} & {\bf MR} &{\bf Hits@10} & {\bf MR}  \\ 
    \hline
    TransE~\cite{TransE} & 0.465 & 357 &  0.501 & 3384 & 0.989 & 1.84 \\
    DistMult~\cite{DistMult} & 0.419 & 254 &  0.49 & 5110 &  0.846 & 5.52 \\
    ComplEx~\cite{trouillon_complex_2016} & 0.428 & 339 &  0.51 & 5261 & 0.967 & 2.59\\
    ConvE~\cite{ConvE} & 0.501 & 244 & 0.52 & 4187 & 0.990&1.51 \\
    RotatE~\cite{RotateE} &  0.533 & 177 & 0.571 & 3340 & -&- \\
    HAKE~\cite{zhang_learning_2019} & 0.542 & -  & 0.582 & -  & - & - \\ 
    KBGAT~\cite{nathani-etal-2019-learning} & \secbest 0.626 & 210 &0.581 & 1940 &  - & - \\ 
    KG-BERT~\cite{KGBERT} & 0.420 & 153 &  0.524 & 97 & 0.990 & 1.47\\ 
    REFE~\cite{chami-etal-2020-low} & 0.541 & - & 0.561 & - & - &- \\
    GAATs~\cite{wang_knowledge_2020} & \best 0.650 & 187 & 0.604 & 1270 & - & -\\ 
    ComplEx-DURA~\cite{NEURIPS2020_f6185f0e} & 0.560 & - & 0.571 & -  & - & -\\ 
    StAR~\cite{wang_structure-augmented_2021} & 0.562 & 117 & 0.732 & 46 & 0.991  &1.49 \\ 
    NePTuNe~\cite{abs-2104-07824} & 0.547 & - & 0.557 & - & - & - \\
    ComplEx-N3-RP~\cite{ChenM0S21} & 0.568 & - & 0.580 & - & \best 0.998 & - \\
     \hdashline
    \method-BERT$_{\rm BASE}$  (\electricblue{\small ours}) & 0.479 & 131 & 0.725 & 55 &  0.991 &\best 1.39\\
    \method-BERT$_{\rm LARGE}$  (\electricblue{\small ours}) &  0.527 &  120 & \secbest 0.769 & \secbest 41 &  0.990 & 1.58 \\
    \method-RoBERTa$_{\rm BASE}$  (\electricblue{\small ours}) & 0.500 & \secbest 116 &  0.737 & 53 &\secbest 0.994 & \secbest1.41 \\
    \method-RoBERTa$_{\rm LARGE}$ (\electricblue{\small ours}) &  0.533 & \best 108 &  \best 0.786 & \best 35 & 0.989 & 1.56 \\
    \bottomrule
    \end{tabular}}
    \caption{{\small Link prediction results on FB15k-237, WN18RR and UMLS.}}
    \label{tab:lp_results}
    \vspace{-0.2in}
\end{table*}

\subsection{Low-Resource Settings}
We additionally test the accuracy of triplet classification in a low-data regime, in particular, when using 5\%, 10\%, 15\%, 20\%, and 30\% of the training data on WN11 and FB13. The results are shown in Figure~\ref{fig:partial-bert}. \method-BERT$_{\rm LARGE}$ consistently outperforms the state-of-the-art KG-BERT. This indicates that \method\ is more data-efficient, as it leverages both semantics and structures in the training data. We also find that \method\ is able to produce competitive results with less training data compared to existing methods even with full training data. \method-BERT$_{\rm LARGE}$ with 5\% training data of WN11 outperforms most of the existing methods using full training data. When using 10\% training data of FB13, \method-BERT$_{\rm LARGE}$ is able to perform comparably with KG-BERT with full training data, and outperforms the remaining methods. This is because \method\ transfers the knowledge about semantics better to the tasks compared to existing approaches without fully leveraging the KG semantics. The results suggest that \method\ is effective in low-resource scenarios.

\subsection{Link Prediction}
\label{sec:link}
Link prediction aims to predict a missing entity given a relation and the other entity, which is evaluated as a ranking problem. We perform link prediction on FB15k-237, WN18RR and UMLS datasets. For each correct triplet \texthrt{(h, r, t)}, either \texthrt{h} or \texthrt{t} is corrupted by replacing it with every other entity in the entity set $E$. These triplets are ranked based on scores produced by Eq.~\ref{eq:score} of \method. The evaluation is under the filtered setting~\cite{TransE}, i.e., removing all the triplets that appear either in the train, dev, or test set. Two common metrics, Mean Rank (MR) and Hits@10 (the proportion of correct entities ranked in the top 10) are used to evaluate the results. A lower MR is better while a higher Hits@10 is better. 
From the results in Table~\ref{tab:lp_results}, we summarize key observations as below.

We find all our methods significantly outperform the compared methods in MR, and reach competitive or better Hits@10.  \method-RoBERTa$_{\rm LARGE}$ performs the best on WN18RR, which outperforms the best compared method StAR by 11 units in MR and 5.4\% in Hits@10. It also delivers the best MR on FB15k-237.  On UMLS, the existing state-of-the-art performance sets a high standard. However, \method-BERT$_{\rm BASE}$ still outperforms others by at least 0.08 unit in MR. The reasons for the improvements are mainly two-fold. (\expandafter{\romannumeral1}) \method\ is able to capture the structural patterns in the existing triplets to predict the missing ones via the structured loss. Compared to KG-BERT, \method\ is able to use the neighboring entities in the KGs for the prediction. (\expandafter{\romannumeral2}) \method\ is able to maintain the semantics of the KGs through semantic embedding to avoid unreasonable triplets with high ranks. For example, if \texthrt{CEO\_Of} holds between two entities, the \texthrt{employee\_Of} also holds, but \texthrt{birth\_Place} does not hold. This is the main reason that \method\ outperforms all structure embedding based methods by a large margin especially in MR. For instance, \method\ significantly outperforms TransE, which shares the similar structured loss with \method. Compare to the improvements made on FB15k-237, \method-RoBERTa$_{\rm LARGE}$ has significantly improved the state-of-the-art results on WN18RR. The main reason leading to such significant improvements is that the pre-trained LMs provide more semantics in the semantic embedding for WordNet as those LMs are trained on textual corpora to capture relationships between words. While WordNet provides the relationships between words, FB15k-237 contains real-world entities and relations, which are less captured by the LMs. 

We also notice that \method\ only produces moderate Hits@10 on FB15k-237. The main reason is that FB15k-237 presents more complex relations between entities compared to other link prediction datasets shown in Table~\ref{tab:statistics}. Therefore, a more complex structured loss is expected for \method\ to gain further improvements. We leave it as one of the future explorations.
Besides, on FB15k-237 and WN18RR, \method-BERT$_{\rm LARGE}$ outperforms \method-BERT$_{\rm BASE}$, and \method-RoBERTa$_{\rm LARGE}$ also outperforms \method-RoBERTa$_{\rm BASE}$. This confirms the recent findings~\cite{LAMA} that larger LMs store more semantic knowledge in the parameters. We expect further improvements when larger LMs are used with \method. On UMLS, we observe slightly different trends. This is mainly because UMLS is a relatively small dataset, thus large models can suffer from overfitting. Overall, RoBERTa improves the BERT pre-training procedure from several perspectives. The improved pre-training procedure enables RoBERTa to generate better performance in many downstream tasks. This suggests that an improved pre-training procedure can enrich the semantics learned in the corresponding LMs.

Both link prediction and triplet classification are core KG completion tasks. The results show that the proposed \method\ generalizes well in KG completion tasks. Different from KG-BERT that designs different models for the tasks, our method does not introduce task-specific parameters or losses for different tasks.

\begin{figure}[htb]
\centering
\subcaptionbox{{\small Semantics.}\label{fig:case_semantic}}{\includegraphics[width=0.25\textwidth]{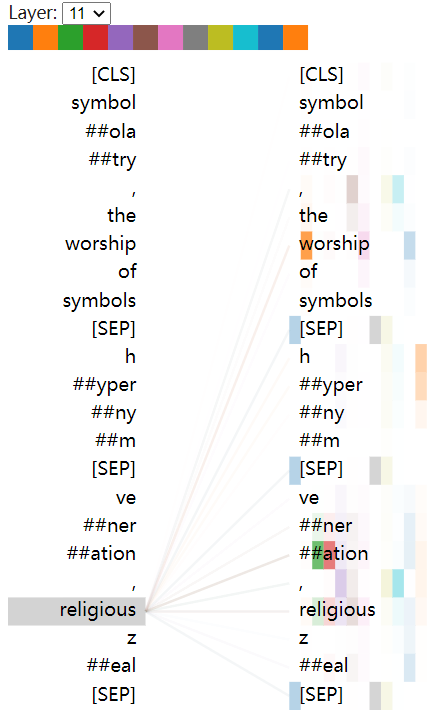}}%
\hspace{0.1in}
\subcaptionbox{{\small Structures.}\label{fig:case_structure}}{\includegraphics[width=0.2\textwidth]{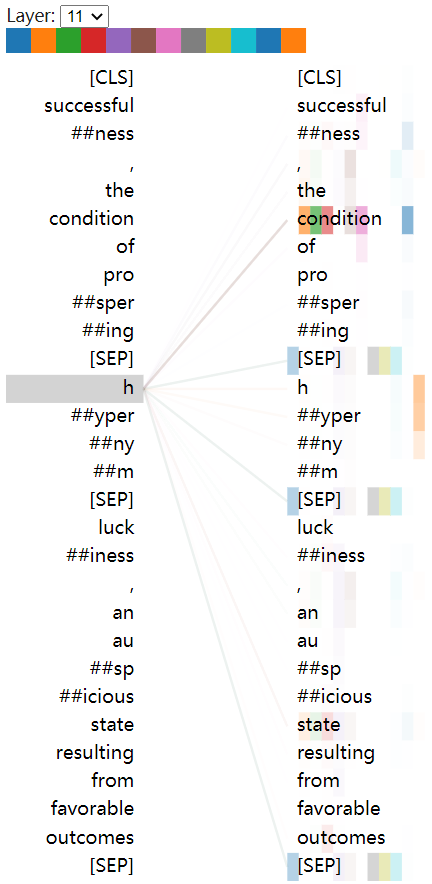}}%
\caption{{\small Illustration of attention weights of the last layer of \method-BERT$_{\rm BASE}$.}}
\label{fig:para}
\end{figure}

\subsection{Case Study}
We show uncurated examples to illustrate why \method~can yield the above results, especially how the parameters of the LMs capture the semantics and structures. As attention layers are basic building blocks of the LMs, we focus on visualizing the attention weights with different input sequences.

We use BertViz~\cite{vig_multiscale_2019} to illustrate the attention weights of the LMs. Given an example of a positive triplet, $h=$ ``symbololatry, the worship of symbols'', $r=$ ``hypernym'', and $t=$ ``veneration, religious zeal'', Figure~\ref{fig:case_semantic} shows the attention weights of the last layer of \method-BERT$_{\rm BASE}$ on WN11. We find that semantically related tokens attend to each other with relatively high scores. For example, ``religious'' attends intensively to ``worship'' and ``veneration''. As in multi-head self attention~\cite{Vaswani_Shazeer_Parmar_Uszkoreit_Jones_Gomez_Kaiser_Polosukhin_2017}, different attention heads in different colors attend to different aspects of the input, the heads are then concatenated to compute the final attention weights. The darker the color, the larger the attention score. This demonstrates that the semantic embedding of \method~is effective in capturing the semantics in the natural language description of the triplets.

We show another positive example with $h=$ ``successfulness, the condition of prospering'', $r=$ ``hypernym'', and $t=$ ``luckiness, an auspicious state resulting from favorable outcomes''. Figure~\ref{fig:case_structure} illustrates the attention weights of the last layer of \method-BERT$_{\rm BASE}$ on WN11. We observe that tokens are highly attended to each other with similar structure roles in the triplet, even though they share fewer semantic similarities. For instance, the attention score between ``hypernym'' and ``condition'' is large. There is also a large attention score between ``hypernym'' and ``state''. This is because both ``condition'' and ``state'' capture the critical structure information of the triplet. The results indicate that the structure embedding of \method~is able to reconstruct the structure information in the semantic embeddings.

\subsection{Error Analysis}
To better understand the limitations of \method, we perform a detailed analysis of the errors. We use triplet classification as an example. We investigate the errors made by \method-BERT$_{\rm BASE}$ on WN11 and summarize the errors based on the relations in Table~\ref{tab:error}. We find most errors are caused by relations that are hard to be distinguished from each other due to their semantic similarities. For example, ``domain topic'' and ``domain region'' are such relations with an unclear semantic boundary.

\begin{table}[htb]
    \centering
    \resizebox{0.85\linewidth}{!}{
    \begin{tabular}{cc}
     \toprule
    {\bf Relation} & {\bf Percentage} (\%) \\
    \hline
    domain topic & 19.8 \\
    domain region & 10.8 \\
    member meronym & 9.1 \\
    has instance & 8.4 \\
    has part & 8.1 \\
    similar to & 7.1 \\
    part of & 6.3 \\
    synset domain topic & 5.5 \\
    type of & 4.6 \\
    member holonym & 3.9 \\
    subordinate instance of & 3.2 \\
    \bottomrule
    \end{tabular}}
    \caption{{\small Analysis of most common errors of \method-BERT$_{\rm BASE}$ categorized by relations on WN11.}}
    \label{tab:error}
\end{table}

\section{Discussion}
\label{sec:dis}
\paragraph{Structure Losses} There are several directions to further improve \method.
\method~uses the probabilistic structured loss based on the score function of TransE, which learns a single representation for every entity and relation in the same embedding space. However, different relationships expect different entity embeddings. We propose to enable an entity to have distinct distributed representations when involved in different relations. For example, a new score function $\Vert \mathbf{h}_r+\mathbf{r}-\mathbf{t}_r \Vert_2^2$ models entities and relations in distinct spaces, and performs the translation between entity embeddings in relation space. The idea is in the same spirit as TransH~\cite{TransH} and TransR~\cite{TransR}. However, a downside of leveraging those losses is that they will bring additional computation overhead. Our method aims to trade off the computation costs and effectiveness. Exploring computation-light methods that involve alternative losses is one of the future investigations.

\paragraph{Pre-trained LMs} We have explored two pre-trained LM families: BERT and RoBERTa. There are three possible directions along this line. First, as indicated in the experimental findings, larger LMs often store more semantics, which can improve the semantic embedding module of \method. We propose to examine larger pre-trained LMs, such as GPT-2~\cite{Radford_Wu_Child_Luan_Amodei_Sutskever_2019}, GPT-3~\citep{Brown_Mann_Ryder}, and Megatron-LM~\citep{shoeybi_megatron-lm:_2020}. Incorporating longer language descriptions (e.g., Wikipedia page) of the entities in the knowledge graphs will provide richer knowledge for improved natural language understanding. Second, the fine-tuning procedure of the deep LMs for KG completion tasks, especially link prediction, is still computationally inefficient. Investigating light LM architectures, such as ALBERT~\citep{lan_albert:_2019}, to speed up the training process, is one of the promising directions. Finally, our proposed method is generally useful for many knowledge-driven downstream NLP tasks (e.g., question answering, factual probing) as well as low-resource NLP tasks. Ensembling our method with autoregressive models (e.g., GPT-2) will enable the method to perform text generation tasks.

\section{Related Work}
\paragraph{Pre-trained LMs}
Pre-trained LMs, such as BERT, have recently been used to obtain state-of-the-art results in many NLP benchmarks~\cite{Devlin_Chang_Lee_Toutanova_2019,Liu_Ott_Goyal_Du_Joshi_Chen_Levy_Lewis_Zettlemoyer_Stoyanov_2019}. These models are usually based on Transformers~\cite{Vaswani_Shazeer_Parmar_Uszkoreit_Jones_Gomez_Kaiser_Polosukhin_2017} and trained on unlabeled text corpora. 
They are used to improve downstream tasks via embedding~\cite{Peters_Neumann_Iyyer_Gardner_Clark_Lee_Zettlemoyer_2018}, fine-tuning~\cite{Radford_2018}, or few-shot learning~\cite{Radford_Wu_Child_Luan_Amodei_Sutskever_2019}. 
Fine-tuning bidirectional Transformers is the most widely used scheme in recent NLP applications, and the approach described in this paper is also based on this scheme. The main difference is that we design a structured loss on top of the LMs aiming to capture structures in natural language.

\paragraph{Knowledge Graph Embedding}
KG embedding aims to map entities and their relations to a continuous vector space. Traditional KG embedding methods represent each entity or each relation with a fixed vector. For any triplet (h,r,t), they use a scoring function $f(\mathbf{h},\mathbf{r},\mathbf{t})$ to model its likelihood. The scoring function of TransE~\cite{TransE} is a negative translational distance. It can be augmented with different geometric transformations such as linear projections~\cite{TransH,TransR} or rotations~\cite{RotateE}. Other models based on bilinear transformations~\cite{DistMult}, and convolutions~\cite{ConvE}, also show promising results on KG completion benchmarks. Our structured loss is motivated by TransE. The main differences are the following. TransE~\cite{TransE} treats the relation as a translation of the embeddings from the head to the tail. Therefore $\mathbf{h} + \mathbf{r} \approx \mathbf{t}$ when \texthrt{(h, r, t)} holds. TransE designs a margin-based ranking loss based on the $l_2$ norm $\Vert \mathbf{h}+\mathbf{r}-\mathbf{t} \Vert_2^2$. The key differences between \method\ and TransE are: (\expandafter{\romannumeral1}) \method\ leverages the natural language semantics in LMs, while TransE does not; (\expandafter{\romannumeral2}) \method\ is a probabilistic structured loss, and is more computationally efficient and data-efficient compared to TransE. The main advantage of the probabilistic loss is that we eliminate the norm calculation that TransE requires to prevent the training process from trivially minimizing its loss by increasing the embeddings of entities or relations. The ranking loss of TransE calculates the loss of some training examples as zeros, which will not contribute to the optimization procedure. Our probabilistic loss makes use of all the training examples. Besides, we introduce corrupted relations in the loss, which provides more flexibility in incorporating the KG structure.

Traditional KG embedding approaches aforementioned regard entities and relations as basic units, without using any extra information. However, studies~\cite{KB_NL_1,KB_NL_2,KB_NL_3} show that a KG model that models the natural language descriptions of entities and relations usually outperforms those methods that only model the structure of knowledge triplets. \citet{LAMA} use LMs as virtual KGs to answer factual questions. ERNIE~\cite{ERNIE} integrates structural KGs into pre-trained models to improve knowledge-driven NLP tasks. By contrast, we aim to combine both the structures and semantics of the KGs via a unified optimization procedure for the task of KG completion. KG-BERT~\cite{KGBERT} models KG completion tasks as sentence classification tasks and solves them by fine-tuning pre-trained LMs. There are several key differences between our \method\ and KG-BERT~\cite{KGBERT}: (\expandafter{\romannumeral1}) \method\ reconstructs the structures of KGs via structure embedding, while KG-BERT does not; (\expandafter{\romannumeral2}) \method\ unifies the link prediction and triplet classification under the same architecture, while KG-BERT designs different architectures for different tasks; (\expandafter{\romannumeral3}) \method\ works with two families of LMs, while KG-BERT only works with BERT$_{\rm BASE}$. \method\ is not particularly designed for BERT, shedding light on understanding the role of semantics in LMs for KG completion.

\section{Conclusion}
We propose a new embedding method that leverages both semantics and structures of the knowledge graphs for the task of knowledge graph completion, and offers additional benefits in low-resource settings. The method maps a knowledge graph triplet to an embedding space via fine-tuning language models, where the forward pass captures semantics and the loss reconstructs structures.
Our method has shown significant improvements on knowledge graph completion benchmarks. 
The implementation has made no modifications to the language model architectures. 
The results suggest that the learned embeddings are generally useful in downstream knowledge-driven applications, and potentially useful for more natural language understanding tasks. We hope our results will foster further research in this direction.

\section{Ethical Considerations}
We hereby acknowledge that all of the co-authors of this work are aware of the provided \textit{ACM Code of Ethics} and honor the code of conduct. The followings give the aspects of both our ethical considerations and our potential impacts to the community.
This work uses pre-trained LMs for knowledge graph completion. The risks and potential misuse of LMs are discussed in \citet{Brown_Mann_Ryder}. There are potential undesirable biases in the datasets, such as unfaithful descriptions from Wikipedia. We do not anticipate the production of harmful outputs after using our model, especially towards vulnerable populations. 

\section{Environmental Considerations}
We use BERT and RoBERTa as our pre-trained LMs. According to the estimation in \citet{strubell-etal-2019-energy}, pre-training a base model costs 1,507 kWh$\cdot$PUE and emits 1,438 lb $CO_2$, while pre-training a large model requires 4 times the resources of a base model. In addition, our fine-tuning takes less than 1\% gradient-steps of the number of steps of pre-training. Therefore, our energy cost and $CO_2$ emissions are relatively small. Besides, the results in the low-resource settings show that our method has better sampling efficiency. This indicates that we can further reduce energy consumption when training with fewer data.

\section*{Acknowledgement}
We would like to thank the anonymous reviewers for their suggestions and comments. This material is in part based upon work supported by Berkeley DeepDrive and Berkeley Artificial Intelligence Research.
% \newpage

%\bibliographystyle{acl_natbib}
%\bibliography{custom}

\appendix
\section{Experimental Setup Details}
We describe additional details of our experimental setup including datasets and comparison methods in this section.

\subsection{Datasets}
\label{apx:data}
We introduce the link prediction and triplet classification datasets as below.

\newcommand{\norm}[1]{\left\lVert#1\right\rVert}
\begin{table*}[htbp]
\centering
\resizebox{1.0\linewidth}{!}{
\begin{tabular}{lcc}
\toprule
\textbf{Method} &  \multicolumn{2}{c}{\textbf{Score Function}} \\
\midrule
TransE & $-\norm{\textbf{h} + \textbf{r} - \textbf{t}}$ & $\textbf{h}, \textbf{r}, \textbf{t} \in \mathbb{R}^k$\\
TransH & $-\left\|\left(\mathbf{h}-\mathbf{w}_{r}^{\top} \mathbf{h} \mathbf{w}_{r}\right)+\mathbf{r}-\left(\mathbf{t}-\mathbf{w}_{r}^{\top} \mathbf{t} \mathbf{w}_{r}\right)\right\|$ & $\mathbf{h}, \mathbf{t}, \mathbf{r}, \mathbf{w}_{r} \in \mathbb{R}^{k}$\\
TransR & $-\norm{\mathbf{M}_r\mathbf{h} + \mathbf{r} - \mathbf{M}_r\mathbf{t}}$ & $\mathbf{h}, \mathbf{t}\in\mathbb{R}^k, \mathbf{M}_r\in\mathbb{R}^{k\times d}$ \\
TransD & $-\left\|\left(\mathbf{w}_{r} \mathbf{w}_{h}^{\top}+\mathbf{I}\right) \mathbf{h}+\mathbf{r}-\left(\mathbf{w}_{r} \mathbf{w}_{t}^{\top}+\mathbf{I}\right) \mathbf{t}\right\|$ & $\mathbf{h}, \mathbf{t}, \mathbf{w}_{h} \mathbf{w}_{t} \in \mathbb{R}^{k}, \mathbf{r}, \mathbf{w}_{r} \in \mathbb{R}^{d}$\\
TransG & $\sum_{i} \pi_{r}^{i} \exp \left(-\frac{\left\|\boldsymbol{\mu}_{h}+\boldsymbol{\mu}_{r}^{i}-\boldsymbol{\mu}_{t}\right\|}{\sigma_{h}^{2}+\sigma_{t}^{2}}\right)$ & $\mathbf{h} \sim \mathcal{N}\left(\boldsymbol{\mu}_{h}, \boldsymbol{\sigma}_{h}^{2} \mathbf{I}\right)$,
$\mathbf{t} \sim \mathcal{N}\left(\boldsymbol{\mu}_{t}, \Sigma_{t}\right)$,
$\boldsymbol{\mu}_{h}, \boldsymbol{\mu}_{t} \in \mathbb{R}^{k}$\\
TranSparse-S & $-\left\|\mathbf{M}_{r}\left(\theta_{r}\right) \mathbf{h}+\mathbf{r}-\mathbf{M}_{r}\left(\theta_{r}\right) \mathbf{t}\right\|_{1 / 2}^{2}$
$-\left\|\mathbf{M}_{r}^{1}\left(\theta_{r}^{1}\right) \mathbf{h}+\mathbf{r}-\mathbf{M}_{r}^{2}\left(\theta_{r}^{2}\right) \mathbf{t}\right\|_{1 / 2}^{2}$ & $\mathbf{h}, \mathbf{t} \in \mathbb{R}^{k}$, $\mathbf{r} \in \mathbb{R}^{d}, \mathbf{M}_{r}\left(\theta_{r}\right) \in \mathbb{R}^{k \times d}$, $\mathbf{M}_{r}^{1}\left(\theta_{r}^{1}\right), \mathbf{M}_{r}^{2}\left(\theta_{r}^{2}\right) \in \mathbb{R}^{k \times d}$\\
DistMult & $ \langle \textbf{r}, \textbf{h}, \textbf{t} \rangle$ & $\textbf{h}, \textbf{r}, \textbf{t} \in \mathbb{R}^k$\\
ConvKB & $\operatorname{concat}(g([\boldsymbol{h}, \boldsymbol{r}, \boldsymbol{t}] * \omega)) \mathbf{w}$ & $\textbf{h}, \textbf{r}, \textbf{t} \in \mathbb{R}^k$ \\
ComplEx & $ \Re(\langle \textbf{r}, \textbf{h}, \overline{\textbf{t}} \rangle)$ & $\textbf{h}, \textbf{r}, \textbf{t} \in \mathbb{C}^k$\\
ConvE & $ \langle \sigma(\mathrm{vec}(\sigma([ \overline{\textbf{r}}, \overline{\textbf{h}}] \ast \boldsymbol{\Omega})) \mathbf{W}), \textbf{t} \rangle$ & $\textbf{h}, \textbf{r}, \textbf{t} \in \mathbb{R}^k$\\
RotatE & $-\norm{\textbf{h} \circ \textbf{r} - \textbf{t}}^2$ & $\textbf{h}, \textbf{r}, \textbf{t} \in \mathbb{C}^k, |r_i| = 1$ \\
REFE & $-\mathrm{arctanh}(\norm{-\langle\mathbf{h}, \mathrm{Ref}(\mathbf{r})\rangle\oplus^c \mathbf{t}})$ & $\textbf{h}, \textbf{r}, \textbf{t} \in \mathbb{R}^k$\\
HAKE & $\text{RotatE}-\norm{\sin((\mathbf{h} + \mathbf{r} - \mathbf{t}) / 2)}_1$ & $\textbf{h}, \textbf{r}, \textbf{t} \in \mathbb{R}^k$ \\
ComplEx-DURA & $\text{ComplEx} - \langle \mathbf{h}, \mathbf{r}\rangle^2 - \norm{\mathbf{t}}^2$ & $\textbf{h}, \textbf{r}, \textbf{t} \in \mathbb{C}^k$ \\
\bottomrule
\end{tabular}}
\caption{The score functions $f_r(\textbf{h}, \textbf{t})$ of shallow structure embedding models for knowledge graph embedding, where $\langle \cdot \rangle$ denotes the generalized dot product, $\circ$ denotes the Hadamard product, $\sigma$ denotes activation function and $\ast$ denotes 2D convolution. $\overline{\ \cdot\ }$ denotes conjugate for complex vectors, and 2D reshaping for real vectors in the ConvE model. $\mathrm{Ref}(\theta)$ denotes the reflection matrix induced by rotation parameters $\theta$. $\oplus^c$ is Möbius addition that provides an analogue to Euclidean addition for hyperbolic space.
\label{tab:structure_methods}}
\end{table*}

\subsubsection{Link Prediction}
\begin{itemize}[leftmargin=*]
\item {\bf FB15k-237}. Freebase is a large collaborative knowledge graph consisting of data composed mainly by its community members. It is an online collection of structured data harvested from many sources, including individual and user-submitted wiki contributions \cite{freebase}. FB15k is a selected subset of Freebase that consists of 14,951 entities and 1,345 relationships \cite{TransE}. FB15K-237 is a variant of FB15K where inverse relations and redundant relations are removed, resulting in 237 relations \cite{text_joint_kb}.

\item {\bf WN18RR}. WordNet is a lexical database of semantic relations between words in English. WN18 \cite{TransE} is a subset of WordNet which
consists of 18 relations and 40,943 entities. WN18RR is created to ensure that the evaluation dataset does not have inverse relations to prevent test leakage \cite{ConvE}.

\item {\bf UMLS}. UMLS semantic network \cite{UMLS} is an upper-level ontology of Unified Medical Language System. The semantic network, through its 135 semantic types, provides a consistent categorization of all concepts represented in the UMLS. The 46 links between the semantic types provide the structure for the network and represent important relationships in the biomedical domain.

\comm{
\item {\bf YAGO3-10}. Yet Another Great Ontology (YAGO) is a knowledge graph that augments WordNet with common knowledge facts extracted from Wikipedia, converting WordNet from a primarily linguistic resource to a common knowledge graph \cite{yago}. YAGO3-10 is a benchmark dataset for knowledge graph completion. It is a subset of YAGO3 (which itself is an extension of YAGO) that contains entities associated with at least ten different relations. Table~\ref{tab:yago} shows the statistics of YAGO3-10 dataset.
}
\end{itemize}

\comm{
\begin{table}[htbp]
    \centering
    \begin{tabular}{ccccc}
    \toprule
        \# {\bf Entity} & \# {\bf Relation} & \# {\bf Train} & \# {\bf Dev} & \# {\bf Test}  \\
        \midrule
        123,182  & 37 & 1,079,040 & 5,000 & 5,000 \\
        \bottomrule
    \end{tabular}
    \caption{Statistics of YAGO3-10 dataset.}
    \label{tab:yago}
\end{table}
}

\subsubsection{Triplet Classification}
\begin{itemize}[leftmargin=*]
\item {\bf WN11 and FB13} are subsets of WordNet and FreeBase respectively for triplet classification, where \citet{KB_NL_1} randomly switch entities from correct testing triplets resulting in a total of doubling the number of test triplets with an equal number of positive and negative examples.
\end{itemize}

\subsection{Comparison Methods}
\label{apx:comp}
We compare \method\ to three types of knowledge graph completion methods: shallow structure embedding, deep structure embedding, and language semantic embedding.\footnote{We refer the readers to \cite{ji2021survey} for a more comprehensive review of the knowledge graph completion methods.}

\subsubsection{Shallow Structure Embedding}
TransE~\cite{TransE}, TransH~\cite{TransH}, TransR~\cite{TransR}, TransD~\cite{ji_knowledge_2015}, TransG~\cite{xiao_transg_2016}, TranSparse-S~\cite{ji_knowledge_2016}, DistMult~\cite{DistMult}, ConvKB~\cite{ConvKB}, ComplEx~\cite{trouillon_complex_2016}, ConvE~\cite{ConvE}, RotatE~\cite{RotateE}, REFE~\cite{chami-etal-2020-low}, HAKE~\cite{zhang_learning_2019}, and ComplEx-DURA~\cite{NEURIPS2020_f6185f0e} are methods based only on the structure of the knowledge graphs. DistMult-HRS~\cite{zhang_knowledge_2018} is an extension of DistMult which is combined with a three-layer hierarchical relation structure (HRS) loss. Each of these methods proposes a scoring function regarding a knowledge triplet, without using the natural language descriptions or names of entities or relations. The scoring functions are shown in Table~\ref{tab:structure_methods}.

\subsubsection{Deep Structure Embedding}
\begin{itemize}[leftmargin=*]
\item {\bf NTN} (Neural Tensor Network)~\cite{KB_NL_1} models entities across multiple dimensions by a bilinear tensor neural layer.

\item {\bf DOLORES}~\cite{wang_dolores:_2018} is based on bi-directional LSTMs and learns deep representations of entities and relations from constructed entity-relation chains.

\item {\bf KBGAT} proposes an attention-based feature embedding that captures both entity and relation features in any given entity’s neighborhood, and additionally encapsulates relation clusters and multi-hop relations \cite{nathani-etal-2019-learning}.

\item {\bf GAATs} integrates an attenuated attention mechanism in a graph neural network to assign different weights in different relation paths and acquire the information from the neighborhoods \cite{wang_knowledge_2020}. 

\item {\bf NePTuNe} takes advantage of both TuckER and NTN by carefully crafted nonlinearities and a shared core tensor intrinsic to the Tucker decomposition \cite{abs-2104-07824}. 

\item {\bf ComplEx-N3-RP} introduces an auxiliary training task to predict relation types as a self-supervised objective. \cite{ChenM0S21}. 
\end{itemize}

\subsubsection{Language Semantic Embedding}
\begin{itemize}[leftmargin=*]
\item {\bf TEKE}~\cite{wang_text-enhanced_2016} takes advantage of the context information in a text corpus. The textual context information is incorporated to expand the semantic structure of the knowledge graph and each relation is enabled to own different representations for different head and tail entities.

\item {\bf AATE}~\cite{KB_NL_6} is a text-enhanced knowledge graph representation learning method, which can represent a relation/entity with different representations in different triples by exploiting additional textual information.

\item {\bf KG-BERT}~\cite{KGBERT} considers triples in knowledge graphs as textual sequences, where each textual sequence is a concatenation of text descriptions of the head entity, the relation, and the tail entity. Then KG-BERT treats the knowledge graph completion task as a text binary classification task, and then solves it by fine-tuning a pre-trained BERT.

\item {\bf StAR}~\cite{wang_structure-augmented_2021} partitions each triplet into two asymmetric parts as in translation-based graph embedding approach, and encodes both parts into contextualized representations by a Siamese-style textual encoder (BERT or RoBERTa).
\end{itemize}

\end{document}